# Cost-effective Machine Learning Inference Offload for Edge Computing

Christian Makaya, Amalendu Iyer, Jonathan Salfity, Madhu Athreya, M Anthony Lewis
HP Labs – AIECL, HP Inc.
Palo Alto, CA, USA

*Abstract*—Computing at the edge is increasingly important since a massive amount of data is generated. This poses challenges in transporting all that data to the remote data centers and cloud, where they can be processed and analyzed. On the other hand, harnessing the edge data is essential for offering data-driven and machine learning-based applications, if the challenges, such as device capabilities, connectivity, and heterogeneity can be mitigated. Machine learning applications are very compute-intensive and require processing of large amount of data. However, edge devices are often resources-constrained, in terms of compute resources, power, storage, and network connectivity. Hence, limiting their potential to run efficiently and accurately state-of-the art deep neural network (DNN) models, which are becoming larger and more complex. This paper proposes a novel offloading mechanism by leveraging installed-base on-premises (edge) computational resources. The proposed mechanism allows the edge devices to offload heavy and compute-intensive workloads to edge nodes instead of using remote cloud. Our offloading mechanism has been prototyped and tested with state-of-the art person and object detection DNN models for mobile robots and video surveillance applications. The performance shows a significant gain compared to cloud-based offloading strategies in terms of accuracy and latency.

*Keywords—Edge computing; machine learning; heavy inference; deep learning; scheduling.*

## I. Introduction

With the exponential adoption of connected edge devices, the applications and services that leverage such devices span across industries such as digital manufacturing, smart home, smart cities, industrial process, autonomous vehicle, and healthcare. The Internet of Things (IoT) is defined as the collection of everyday machines, devices, and smart objects that are now embedded with sensors and/or actuators and that can communicate over the Internet. One of the biggest challenges posed by such adoption is the exponential growth of data generated by these edge devices. It is estimated that the data growth is outpacing network bandwidth improvements by a factor of two. While insights extracted from such data are extremely valuable to businesses and users alike, the trends above indicate that having to transport such high volume of data over the network will be increasingly challenging. In fact, it is estimated that more than 80% of the data remains untapped – not analyzed, stored, or transported, due to the network connectivity, security and privacy, and real-time low latency constraints [1].

On the other hand, the onboard computational resources of edge devices such as IoT devices, drones, smart speaker, video surveillance camera, and mobile robots need to be kept relatively small to reduce power usage and manufacturing cost. Lowering the cost and extended power autonomy are required for mass adoption and higher penetration rate for consumers and industrial usage as we are witnessing. With the progresses and advances in machine learning (ML) and deep learning (DL), there is an increasing adoption of ML-based applications and services in various industries. However, ML-driven applications are very compute-intensive and require the processing of large amount of data. The size of ML models (e.g., perception and language models) are becoming very large, and often can't fit in memory of edge devices. Such models cannot be deployed in these devices, leading to the need of distributing and offloading the computation.

To achieve the full potential of ML, the workloads are run on compute nodes with discrete or dedicated GPUs. For the remaining of the paper, GPU refers to a discrete or dedicated GPU. However, edge devices don't necessarily have such high-end capability to efficiently run these ML models. The state-of-the-art paradigms for analyzing data generated at the edge are two-tiered: a centralized tier of cloud services and a distributed collection of edge devices connected to the cloud. In such paradigms, most of the intensive computation such as big data analytics and machine learning computation occur in the cloud tier, requiring to send huge amount of data to the cloud data centers. The bandwidth limitation, higher latency, and the cost of transporting the data to the cloud are well understood challenges in the edge computing literature. Although, offloading computation to the cloud reduces the onboard computing requirements of edge devices, it results in higher latency that could severely degrade the performance of the applications and have potential privacy and security threats. For example, sending HD video and LiDAR data points from multiple edge devices to a remote cloud could be very expensive and might lead to network congestion, higher latency which is prohibitive for real-time applications such as object detection for video surveillance and autonomous vehicles. Moreover, if a network is congested, there is a risk of information loss and failure.

Furthermore, getting access to GPU instances, whether using purchased cards or through cloud providers (e.g., AWS, Azure) might be very expensive since running ML workloads (e.g., training, inference) might take hours or days. Several analyses of the on-premises cluster utilization show that resources are often under-utilized and idle most of time. The spare GPUs or computational resources that are not used can benefit to other ML tasks, if they are made available. This helps to reduce the cost of GPU and ML accelerators over time instead of relying on expensive cloud services. Hence, the ML jobs can be offloaded to the on-premises computational resources, residing closer to the edge. This results to the minimization of communication overhead, preserving privacy, minimizing latency, minimizing power consumption of edge devices while guaranteeing higher performance of the ML models.

Due to the heterogeneity of edge computational resources and the requirements of the applications, offloading the computation for ML workloads is a challenging task. In fact, it requires addressing the question: *how to efficiently manage and orchestrate computational resources for ML training and inference?* To address this challenge, this paper proposes a scheduling and placement algorithm. With the characteristics of ML jobs, considering time-based (temporal) or size-based (spatial) scheduling is not enough. In fact, the duration of ML jobs is often unknown or can't be predicted even when smooth loss function is considered. On the other hand, considering only the size of the jobs (e.g., number of GPUs) is not efficient since the duration of the jobs is ignored. Moreover, ML jobs have the unique characteristic of all-or-nothing, meaning, the computational resources required for an ML job should be allocated at the same time, known as gang scheduling [10]. Based on these observations, we propose a two-dimensional (spatial and temporal) scheduling mechanism.

Shortest-remaining-service-first (SRSF) [8] is an example of scheduling strategy that considers both spatial and temporal dimensions. But, in SRSF, the remaining service is the multiplication of the number of GPUs and the job's remaining time. As mentioned above, the duration of ML jobs is unknown, thus the remaining time. SRSF can't be applied efficiently for ML jobs. Furthermore, since the edge environments are highly constrained, there is a need for preemptive mechanism for the scheduling. An example of preemptive scheduling mechanism is the shortest-remaining-time-first (SRTF). Least-Attained-Service (LAS) [practice] is a preemptive scheduling policy that doesn't require prior knowledge of the size of the jobs. It also estimates the remaining service time of a job based on the service it has received so far [practice].

This paper proposes a novel offloading mechanism for compute-intensive ML workloads such as ML-based video surveillance, to address the limitations of edge devices in terms of computational resources, storage, and network bandwidth as well as to avoid highly expensive resources from the cloud. The proposed solution automates the orchestration and deployment of ML-based distributed edge applications while guaranteeing efficient usage of resources, scalability, and fault-tolerance. Specifically, it enables resource-aware deployment of ML edge applications.

The remaining of this paper is organized as follows: Section II describes the related work; Section III presents the proposed ML tasks offloading mechanism and discuss issues for compute-intensive workloads in edge computing. The prototype and the benchmark of ML tasks offloading mechanism for start-of-the art DNN models are described in Section IV while Section V concludes the paper.

## II. BACKGROUND AND RELATED WORK

The edge computing approaches such as cloudlets propose to localize as much as possible computation, applications, data storage, and other computing services on the edge tier. Although these approaches are a step in the right direction, they suffer from the limitations of the two-tier paradigm. In the two-tier architecture, most of the edge devices are deployed in private networks and are owned by independent entities, e.g., homes, hospitals, and industrial plants. Furthermore, except for some edge devices (e.g., sensors, actuators, drones, camera, smart speaker) which are resources-constrained, there are edge gateways (e.g., workstations, IoT hub devices, notebooks, high-end drones, smartphones, tablets) that have significant computational, storage, and communication resources. Two-tier approaches don't fully take advantage of the edge gateways for computation needs instead focus on limitless computation in the cloud.

Moreover, the edge environments are highly dynamic due to the mobility of the devices (e.g., robots, drones, autonomous vehicles) as well as the changes in the available resources. The state-of-the-art is missing resources- and context-aware mechanisms for the deployment of analytics such that resources can be allocated optimally. Lastly, in the current deployment approaches, the application topology as well as the binding to physical target devices is fixed at the time of the deployment. Given the dynamic nature of edge computing environments, this limits the class of applications that can be supported. The drawbacks of the two-tier architecture have motivated the proposal of the three-tier architecture as illustrated in Fig. 1, such as ETSI's multi-access edge computing (MEC) framework [2].

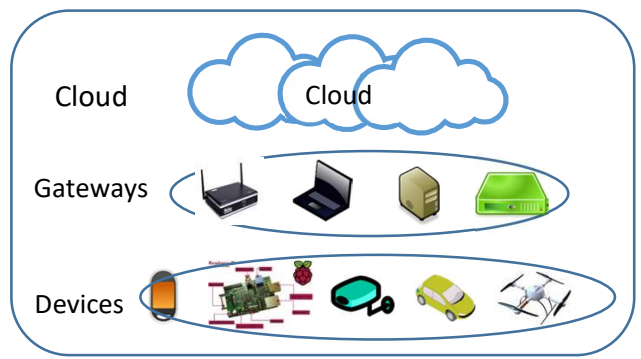

Fig. 1. Three-tier architeture for Edge Computing.

With the three-tier architecture, machine learning functionalities and services can be deployed or offloaded at the edge gateway where raw data processing (aggregations, filtering) can be done and possibly generate actionable events. The edge gateway adds control and improves management closer to the edge rather than sending the raw data to the cloud (which leads to a higher bandwidth consumption, increased latency). Only the insights and possibly filtered data and metadata are sent to the cloud for reporting and further analysis. Such proximity to the edge is particularly useful for real-time and low latency applications. However, 3-tier architecture still rely on the cloud resources to provide enhanced performance. This dependency to the remote cloud might lead to similar issues such as privacy preserving, latency, etc. as observed in 2-tier architecture.

Limited works have been done for ML-based applications while trying to leverage on-premises computational resources. The existing solutions have been designed for high-end infrastructures and devices. CloneCloud [3] is a code-offloading system that automatically partition applications and execution runtime and offload intensive computational tasks from mobile devices onto a device's clone running in the cloud. By offloading application's tasks to the cloud, CloneCloud cannot guarantee the context accuracy for applications. Furthermore, CloneCloud doesn't address specific requirements of ML workloads such as gang scheduling for ML training.

In [4], a robot offloading problem is formulated as a sequential decision-making problem for robots and a deep reinforcement learning solution is proposed. The lack of support for gang scheduling and preemption leads to head-of-line and fragmentation issues as well as ML jobs failure. Several characteristics should be addressed for ML-based applications offloading at the edge. For example, an efficient resources management, isolation, scalability, and fault tolerance should be enabled due to the specific requirements of the ML-based edge applications in terms of low latency, communication overhead, computational efficiency, energy efficiency, and privacy. Our paper addresses a selected set of these requirements by proposing an efficient ML offloading for distributed edge computing environments.

Cloud services, like AWS, provide inference services, such as Amazon Polly, Lex, and Rekognition. The suite of ML inference services enables devices to send data to the cloud for processing and receive results. However, these requests are subject to the congestion and latency of internet connection, often in the order of a few seconds. In addition, the cost of streaming dense information, such as images and HD video, is not energy efficient nor cost effective. For applications which are in a tight control loop, such as on a mobile robot using ML inference results for the navigation stack, conversing, and other processes, the latency leads to a higher probability of failure, as seen in our results, Section IV-B.

In [5] a cluster scheduling framework that uses domain-specific knowledge to improve latency and efficiency of training deep learning models called Gandiva is proposed. In Gandiva, DNN jobs are scheduled based on time-sharing algorithms, which were designed for isolation by using fair sharing, instead of minimizing the average job completion time (JCT). Gandiva also provides an efficient mechanism for preempting DNN jobs. To address shortcomings of Gandiva, such as the limited improvement in terms of average job completion time, [6] proposed another cluster management framework tailored for distributed deep learning jobs called Tiresias. It minimizes the average JCT with partial or no prior information. Tiresias introduces two scheduling algorithms: Discretized 2D-LAS and Discretized 2D-Gittins Index.

III. ML OFFLOADING SCHEME

In order to consider machine learning enabled edge applications there is a need for an efficient framework and scheduling approaches that considers various resources and edge attributes such as data locality, geo-location of ML nodes, privacy, and connectivity. Such framework should address the requirements of ML jobs while guaranteeing accuracy of the models and performance such as minimized average job completion time, high utilization of resources, compute and communication efficiency, and starvation avoidance. Furthermore, availability is also a key requirement. In fact, some level of ML reliability should be considered, for example by using checkpoints and redundancy.

Due to the heterogeneity of edge computing, it is crucial to have agnostic framework for edge devices without requiring changes in the edge devices. In other words, the framework should provide a generic set of capabilities for edge applications deployment and should allocate the resources pool to the distributed applications. The resources are of different types: compute (e.g., CPU, memory, I/O, disk, GPU), network (e.g., bandwidth), location and characteristics of data, and hardware requirements. To reduce the communication overhead, it is important to achieve better locality or affinity while scheduling the ML jobs since the consolidation of ML jobs has an impact on the performance of training.

Devices such as smartphones, notebooks, and workstations can adopt the roles of edge devices as well as edge gateways, both at the same time. Among the edge gateways, one acts as the main controller node, herein called distributed ML server, while the other edge nodes belonging to edge gateway category act as ML offloading nodes, called ML workers, in the remaining of this paper. The key differentiating capabilities of the proposed solution in this paper are:

- Resource isolation, scalability, and fault tolerance while supporting various ML edge applications deployment requests across edge devices.
- Dynamic computational resources discovery.
- ML jobs scheduling: based on two-dimensional (spatial--temporal) least attained service (ST-LAS).

Our paper shares similar concepts with Tiresias [6]. However, instead of assuming a large and high-end cluster, like in Tiresias, our solution focuses on heterogeneous and constrained-edge environment. Furthermore, our solution has been designed for real-time edge intelligence applications.

*A. Challenges for ML Edge Applications Management*

Several edge applications require processing data closer to the edge devices (either on the device or gateway). Most of the state-of the art schedulers for big data analytics do not provide primitives for data locality, only the host resources (e.g., CPU, memory, disk, GPU, application port number). This may cause processing to occur far from the edge devices which would translate in high energy expenditure and transmission cost of the data across the network. For training, the ML workloads scheduling requires that workers to be scheduled simultaneously, also known has gang scheduling. In case of inference, this requirement is applicable if the inference task is split across worker nodes and frame inter-leaved inferencing might likely need some complicated synchronization and reassembly. Due to the potential lack of resources availability at a given time slot, a scheduling decision can be delayed hence, the scheduler can wait until data locality and affinity constraints are satisfied, leading to increased average job completion time. On the other hand, naive colocation of ML jobs on the same ML worker may lead to interference due to the contention for shared resources [5] and exclusive access to GPU. An efficient consolidation technique for example based on ML models profiling can improve the performance.

Since the edge is heterogeneous and various applications can be using simultaneously the resources, it is important to provide resources isolation while scheduling multiple tasks. The ML server performs the isolation of the resources when deciding to assign ML jobs to the ML workers. This metric indicates how much a new ML task or job will suffer given the current state of the ML worker. Thus, instead of isolating tasks, a metric which is a function of running tasks is computed and this metric is communicated back to the ML server. The metric will show whether the application can be executed in the ML worker or not. The metric can also be used to provide preference of one ML worker over another. The metric is computed based on continuous monitoring of resources on ML workers. This metric can be defined as the level of intrusiveness of ML jobs on ML worker.

*B. ML Inference Offloading*

This paper proposes a novel ML tasks offloading mechanism for compute-intensive ML workloads such as machine learning inference, to address the limitations of edge devices in terms of computational resources, storage, and network bandwidth as well as to avoid expensive resources from the cloud. The proposed solution automates the orchestration and deployment of ML-based distributed edge applications while guaranteeing efficient usage of resources, scalability, and fault-tolerance. Specifically, it enables resource-aware deployment of ML edge applications. Spatio-temporal is the most widely used contexts for edge resource management. Our solution provides the interfaces and mechanisms for such contextualization of the resource management for ML tasks deployment.

To avoid a single point of failure for the ML server, Apache Zookeeper, a distributed consensus system, is used for reliable election of the leading ML server. The ML server automatically discovers ML workers and they report their available resources (e.g., memory, CPU, ports, disk, GPU) to the ML server, which in turn aggregate such reports. Another key contribution of our solution is the resilient management of tasks failures. The focus of our solution is to enable deployment of advanced ML jobs, mainly heavy deep learning neural networks (DNN) models for inference at the edge. In fact, DNN models for object detection are becoming crucial in various applications.

Upon submission of job request, the ML server (scheduler) validates the application's requirements such as computational resources, latency threshold, data locality, affinity, runtime dependencies, and network connectivity against available resources. The ML server performs the matching of requirements and application dependencies with the available resources from the pool of ML workers. The job is assigned a service priority based on its attained service, which depends on the number of required computational resources (e.g., number of GPUs) and the executed time (i.e., the time the job has been running so far). The service priority of a job is initialized to 0 when the job arrives. The jobs start with highest priority and the priorities decrease as the jobs receive more service. The lower the value of the service priority, the higher the priority, and the closer to the head of the queue the job is inserted [9]. By computing the priority of the job, our algorithm ST-LAS can insert the job at the appropriate position in the queue. A normalized function is used to aggregate computational resources (e.g., number of GPUs, memory, CPU). To avoid jobs starvation and enable preemption, we use discretized multi-level feedback queue (MLFQ) [7]. Instead of having $K$ logical queues, in practice two queues are enough to classify the jobs based on their priority [9]. For the remaining of the paper, we will focus on MLFQ based on 2 queues.

A discretized MLFQ has the advantage of reducing the number of preemptions. Upon arrival, a job is placed in the highest priority queue $Q_1$. An ML job is demoted to $Q_2$ if its attained service crosses a given threshold. The jobs with similar priority (belonging to a given interval) are placed in the same queue. To avoid starvation, a job that has been waiting for longer that a wait threshold is promoted to a high priority queue, i.e., $Q_2$. LAS with threshold (*LAS-threshold*) has been proved to be efficient [9]. By computing the priority of the job, our algorithm ST-LAS can insert the job at the appropriate position in the queue. Furthermore, the ML server can dynamically move tasks across ML workers. Upon the decision to migrate the job, the ML worker checkpoints the most recent model to a persistent storage. When the ML job resumes, all the ML workers reload the checkpoint model file. With the optimized resources management, the proposed solution allows ML-based edge applications to scale automatically based on the application and business logic.

The candidate nodes are selected among ML workers with minimum load and satisfying the ML jobs requirements. Since the tensors size of DNN models cause network imbalance and contention, a consolidated placement is performed for highly skewed models. Such consolidation reduces the communication overhead across ML workers. A model skewness factor represents the parameter size of the model across layers. The model is profiled based on the skewness and the network traffic.

IV. IMPLEMENTATION AND PROTOTYPE

*A. Prototype Description*

We have implemented our solution on a real system prototype. The state-of-the art ML models for object and person detection from TensorFlow Model Garden [11] have been used to assess the performance of the proposed mechanisms and prototype. The DNN models are deployed using TensorFlow Serving (TFS) [12] running in docker containers, to enable resource isolation and mitigate exclusive access to GPU requirement of ML jobs. With TFS, we can deploy ML models for inference across multiple compute nodes. Raspberry Pi 3 [13] cameras are mounted on TurtleBot3 (TB3) Burger [14] robot composed of a Raspberry Pi 3 (RPi3), OpenCR, 360º

LiDAR scanner. The TB3 with a RPi3 and camera was commanded by ROS (Robot Operating System) [15] and the motors were controlled by the OpenCR board. The camera onboard the TB3 streamed images through ROS towards TFS model server running on ML workers. For the comparison with cloud, we deployed AWS EC2 instance p3.2xlarge [17] in us-west-eb availability zone.

The robots are deployed to perform person and object detection for person tracking and map labeling applications. However, due to the low-cost hardware, there is limited on-board computational capabilities and they cannot run heavy and compute-intensive ML inference. The ML inferences are then offloaded towards on-premises ML workers (edge nodes). TFS containers are deployed on GPU-powered compute nodes with shared GPU capability. Both gRPC and REST API modes are enabled on the TFS containers, but for the tests, only REST API was used. The prototype provides a ML preprocessing pipeline module used for image resizing, standardization or normalization. Fig. 2 shows a on-premises prototype setup.

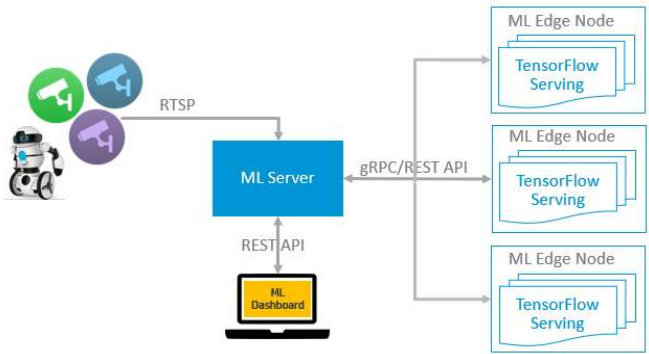

Fig. 2. Prototype architecture for the object detection use case.

## B. Performance Results

Table 1 shows the cost of running ML inference in the cloud compare to the on-premises deployment on workstations for object detection DNN models. For estimating electricity cost we assumed that the GPU enabled workstation consumes 500 watt-hour and 0.2$/Kilowatt-hour for commercial usage in California per the US Energy Information Administration [20].

Table 1: Cost comparison – Cloud vs On-premises.

| Compute Type | Compute Details | Model Provided | Unit Price | Cost for 1 video stream month |
|---|---|---|---|---|
| AWS p3.2xlarge | Cloud instance with datacenter class V100 GPU (14 TFLOPS) | Open Source or custom | $3.06/hour | $2277 |
| AWS Rekognition Video | Unknown | Yes | $0.1/min | $4464 |
| AWS Rekognition Image | Unknown | Yes | $0.4/1000 images | $1071 [1 FPS] |
| cloGCP Vision Object | Unknown | Yes | $1.5/1000 images | $4017 [1 FPS] |
| On-premise workstation with GPU | Workstation with RTX2080 (10TFLOPS) | Open source or custom | | $5000 (One time) and $75 electricity cost/month |

For real-time applications, inference latency (prediction speed) is critical for applications requirements and user experience. It is expected the results to be returned as fast as possible. The prediction speed has a direct impact on the cost of serving. In fact, faster prediction (i.e., lower latency) means higher prediction throughput (i.e., number of predictions per unit of time) on a given compute node, leading to overall reduced cost. Table 2 shows the performance, prediction speed, of different DNN models for edge applications compared to running in a cloud, such as AWS p3.2xlarge instance. A HD video stream is sent from the mobile robot and a snapshot 640x480 JPEG image snapshot of size 192KB was used as payload for calculating the round-trip time. The inference result is published in the ROS pub/sub messaging queue. There is always a tradeoff between accuracy, latency, and compute time. A model might be most suitable for constrained-resources edge devices since it has fewer parameters and operation compare to more accurate models such as Mask R-CNN. Since the proposed framework enables ML inference offloading, using more accurate models gives improved performance in terms of accuracy and processing time. Hence, they might be favored for edge applications.

Table 2: DNN models performance comparison.

| DNN | Model Size (MB) | GPU Inference (ms) | | CPU Inference (ms) | |
|---|---|---|---|---|---|
| | | Cloud | OnPrem | Cloud | OnPrem |
| Faster RCNN ResNet 101 | 196.5 | 1517 | 114 | 1615 | 676 |
| SSD MobileNet V1 | 29.1 | 340 | 77 | 459 | 54 |
| SSD MobileNet V2 | 69.7 | 354 | 75 | 719 | 65 |
| SSD Inception V2 | 102 | 366 | 80 | 624 | 69 |
| Faster RCNN Inception V2 | 57.2 | 619 | 89 | 726 | 221 |
| Mask RCNN Inception V2 | 67.1 | 789 | 97 | 801 | 321 |
| DeepLab V3 | 23 | 925 | 70 | 1371 | 272 |
| DeepLab V3 Xception | 447 | 3124 | 147 | 3061 | 1585 |

Fig. 3 shows the roundtrip inference latency when GPU is not enabled. As we can see, the latency for Cloud settings is higher. The proposed ML offloading at the edge (Edge Distributed AI) outperforms the cloud scenario. When GPU is enabled on ML workers, the latency for Edge Distributed AI (Edge DAI) is further reduced compare to the cloud offloading mechanism, Fig. 4.

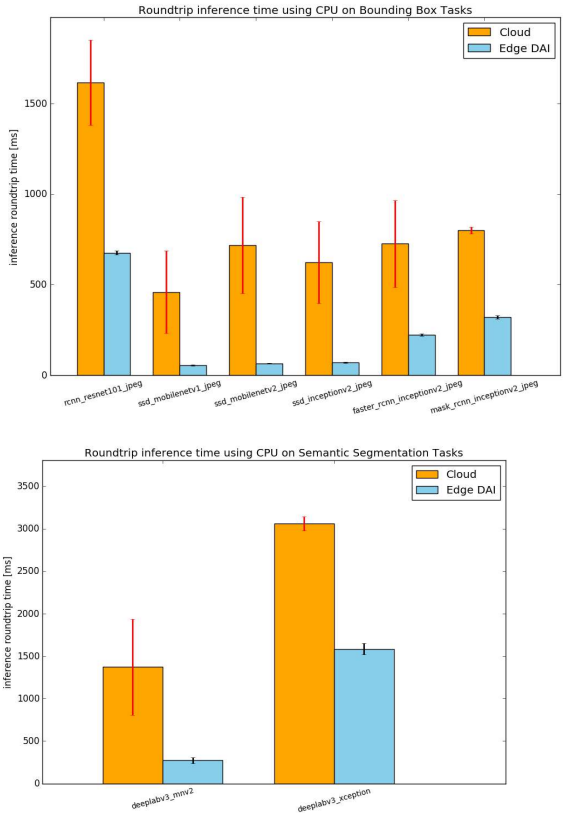

Fig.3. Inference latency for non-GPU-enabled.

The roundtrip inference time is calculated as the total time to send the payload to the processing nodes (ML workers) and receive the inference results. The round-trip time reported in Table 2 is an average 100 requests. All models in Table 2 are object detection models except the DeepLab-V3 and DeepLab Xception model, which is a segmentation model based on MobileNet-V2. Unlike the object detect models that send bounding box co-ordinates the segmentation model sends back a segmentation mask for each pixel. As a result, the round-trip time is dominated by time required to transmit information instead of model execution time. The model size reported in Table 2 is the memory footprint of the frozen graph format from TensorFlow. In the case of offloading the inference into edge ML workers, the roundtrip inference time is significantly lower as well as the acquisition cost. Compared to AWS EC2 instance that requires new acquisition, edge ML workers are leveraged from prior acquisition. The roundtrip inference time may be affected by several factors, which may include compute resources availability in the cloud and on-premises, network upload and download speeds.

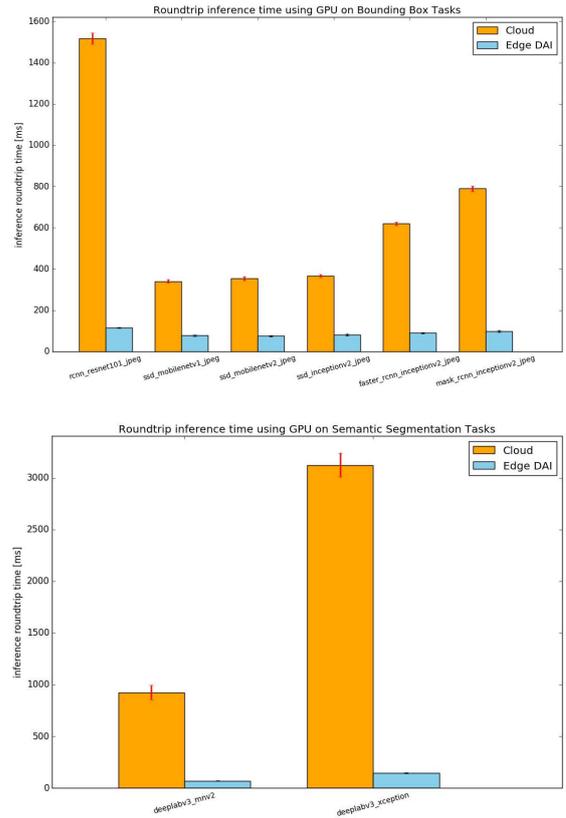

Fig.4. Inference latency for GPU-enabled.

In the context of robotics, the roundtrip inference time is important as the results directly influence the robot's high and low-level control policy. The inference delays can have significant and even catastrophic consequences due to the common real-time robotics applications. When the robot's control policy input signals become slower than the time to take a required action, the agent's behavior and actions may become unpredictable and unstable.

For example, the TB3 robot used in this experiment is set up to track people walking in an office environment and update a semantic map via semantic segmentation. The motor controller comprised of a feedforward and PID architecture implemented by the onboard OpenCR run at 100 Hz with very small motor lag, allowing the hardware actuation architecture to have plenty of margin for tracking and following people. The slowest part of this control system is the reference signal generation. For person tracking, the reference signal is generated by the labeled pixels returned from the inference request.

The average walking speed of a person is around 1.5 m/s. Using the inference requests to the cloud for GPU-enabled scenario, the inference response using DeepLab-V3 comes back to the robot on average at around 779 ms with a relatively large variance. In fact, in close proximity for example, a robot that looks at a person moving at the far end of the retail store has much more time than, when a person is hurrying across. In 779 ms, the desired target may have moved up to more than half a meter, which in our hardware configuration, is beyond the robot's camera vision frame, i.e. the person is relatively close the robot. The result is a failure and potential collision, as the robot is not allowed to complete its primary objective of person tracking.

When the robot's inference is offloaded using the proposed approach in this paper, the inference request becomes not only 10x faster but also more consistent, on average at around 72 ms with a smaller variance. This improved performance allows the person to stay in the robot's field of view and therefore allow the robot to complete the objective.

## V. CONCLUSION

In this paper, we have proposed a novel ML tasks offloading scheme at the edge by leveraging on-premises installed-based workstations instead of sending raw data to a remote cloud for processing. A two-dimensional (spatial and temporal) scheduling is used to orchestrate and manage the workload. Our solution is suitable for resource-constrained edge devices, that have small onboard computation power. The proposed solution shows that, large ML models, such as state-of-the art object detection and video surveillance can be deployed on ML workers closer to the edge for efficient performance while avoiding expensive cost of cloud resources. In fact, the solution automates the orchestration and deployment of distributed edge ML-based applications while guaranteeing efficient usage of resources, scalability, and fault-tolerance. Resources abstraction is introduced in which the ML server acts as the controller for the ML edge nodes. A prototype has been presented and a use case for object detection DNNs models has been described. Moreover, issues and challenges for deploying ML-based edge applications have been presented.